\documentclass[10pt,twocolumn,letterpaper]{article}

\usepackage{cvpr} 

\usepackage{caption}

\definecolor{cvprblue}{rgb}{0.21,0.49,0.74}
\usepackage[pagebackref,breaklinks,colorlinks,allcolors=cvprblue]{hyperref}

\title{Generalizable NGP-SR: Generalizable Neural Radiance Fields Super-Resolution via Neural Graph Primitives}

\author{
Wanqi Yuan\\
Clemson University\\
{\tt\small wanqiy@clemson.edu}
\and
Omkar Sharad Mayekar\\
Clemson University\\
{\tt\small omayeka@clemson.edu}
\and
 Connor Pennington\\
Clemson University\\
{\tt\small cjpenni@clemson.edu}
\and
Nianyi Li\\
Clemson University\\
{\tt\small nianyil@clemson.edu}
}

\usepackage{multirow}
\usepackage{xcolor}

\begin{document}
\maketitle
\begin{abstract}


Neural Radiance Fields (NeRF) achieve photorealistic novel view synthesis but become costly when high-resolution (HR) rendering is required, as HR outputs demand dense sampling and higher-capacity models. Moreover, naively super-resolving per-view renderings in 2D often breaks multi-view consistency. We propose Generalizable NGP-SR, a 3D-aware super-resolution framework that reconstructs an HR radiance field directly from low-resolution (LR) posed images. Built on Neural Graphics Primitives (NGP), NGP-SR conditions radiance prediction on 3D coordinates and learned local texture tokens, enabling recovery of high-frequency details within the radiance field and producing view-consistent HR novel views without external HR references or post-hoc 2D upsampling. Importantly, our model is generalizable: once trained, it can be applied to unseen scenes and rendered from novel viewpoints without per-scene optimization. Experiments on multiple datasets show that NGP-SR consistently improves both reconstruction quality and runtime efficiency over prior NeRF-based super-resolution methods, offering a practical solution for scalable high-resolution novel view synthesis. Our project page and code are available at \url{https://wanqiyuan.github.io/Generalizable-NGP-SR/}.

\end{abstract}    
\section{Introduction}
\label{sec:intro}


Neural Radiance Fields (NeRF) \cite{mildenhall2020nerf} have significantly advanced 3D scene reconstruction and novel view synthesis, enabling photorealistic rendering from sparse 2D images. Follow-up works extended NeRF to support anti-aliased multiscale representations \cite{barron2021mip,yu2024mip,zhang2025lookcloserfrequencyawareradiancefield},
dynamic scene \cite{yan2023nerfds,li2021neuralsceneflowfields,sun2024dyblurf},  
real-time rendering \cite{yu2021plenoctrees,duckworth2024smerf,gu2023ue4nerf},
and inverse rendering\cite{wu2025pbrnerfinverserenderingphysicsbased}. 
Despite these advances, NeRF’s rendering quality is highly dependent on high-resolution (HR) inputs, and producing HR images is computationally expensive due to dense sampling and large network capacity. A natural solution is super-resolution (SR), which reconstructs high-frequency details from low-resolution (LR) inputs. However, naive 2D SR applied to NeRF renderings often breaks multi-view consistency, leading to artifacts in novel views.
Existing NeRF-based SR methods attempt to mitigate this but face significant limitations. Optimization-based methods~\cite{wang2022nerf,huang2023refsr,han2024super} often rely on HR reference images, limiting generalization, or struggle to maintain fine-detail consistency. While recent diffusion-based approaches~\cite{lee2024disr,chen2025bridging} improve perceptual fidelity, they introduce stochastic geometric hallucinations and incur high computational costs, rendering them impractical for real-time applications. Similarly, adaptations of 3D Gaussian Splatting (3DGS) for SR~\cite{shen2024supergaussian,wan2025s2gaussiansparseviewsuperresolution3d} suffer from the overhead of discrete primitives and scalability issues at large magnification factors.


Furthermore, while generalizable NeRF frameworks~\cite{mvsgaussian,gefu,fang2025depth} have emerged to enable fast reconstruction across diverse scenes, they typically prioritize geometry over texture fidelity, resulting in blurred outputs when scaling to high resolutions. This highlights a critical gap: the need for a framework that combines \textit{generalization} with \textit{3D-consistent super-resolution} without the massive overhead of generative foundation models.

In this work, we propose \textbf{NGP-SR}, a novel framework that reconstructs high-resolution radiance fields directly from LR inputs and camera poses using Neural Graphics Primitives (NGP). Unlike prior methods that rely on 2D refinement or expensive generative priors, NGP-SR learns high-frequency details directly in 3D space. We achieve this by conditioning the radiance field on \textbf{texture tokens} extracted from multi-resolution local feature patches. 
These tokens provide patch-level context and effectively induce implicit spatial offsets during radiance evaluation, resembling a multi-sampling behavior that helps recover fine structures while maintaining multi-view coherence. 
By integrating these token-augmented features within NGP’s efficient hash encoding, our method synthesizes high-fidelity HR renderings with strong generalization to unseen scenes.
Our key contributions are:
\begin{itemize}
    \item We propose \textbf{NGP-SR}, a 3D-aware super-resolution framework built on \textbf{feature-enhanced Neural Graphics Primitives}, which conditions radiance prediction jointly on spatial coordinates and multi-resolution \textbf{local patch texture tokens}.

    \item We demonstrate that token-conditioned radiance learning can recover high-frequency details while promoting \textbf{multi-view consistency} from only LR inputs and camera poses, without requiring HR reference images.

    \item We show strong \textbf{generalization to unseen scenes}, enabling direct HR synthesis for novel viewpoints from LR observations, with improved quality and efficiency over prior NeRF-based SR methods.

\end{itemize}

\section{Related Work}
\label{sec:related}


\paragraph{Neural Radiance Fields (NeRF).}
Since the introduction of NeRF~\cite{mildenhall2020nerf}, significant efforts have focused on improving its efficiency and applicability. Works such as Mip-NeRF~\cite{barron2021mip}, Zip-NeRF~\cite{barron2023zip}, and others~\cite{hu2023mipvog,nam2023mipgrid,liu2024ripnerf,hu2023trimiprf,yu2024mip,zhang2025lookcloserfrequencyawareradiancefield} tackled anti-aliasing and scale issues. Others extended NeRF to dynamic scenes~\cite{pumarola2021dnerf,yan2023nerfds,luo2024d2rf,sun2024dyblurf} and real-time rendering~\cite{yu2021plenoctrees,rojas2023rerend,duckworth2024smerf,gu2023ue4nerf,hedman2021bakingnerf}. Instant-NGP~\cite{mueller2022instant} introduced multi-resolution hash encoding for near-instant reconstruction, while 3D Gaussian Splatting (3DGS)~\cite{kerbl20233d} utilized explicit primitives for real-time speeds.
To avoid per-scene optimization, generalizable frameworks emerged. MVSGaussian~\cite{mvsgaussian} and GeFU~\cite{gefu} leverage multi-view stereo and geometry-aware fusion, respectively, while GDB-NeRF~\cite{fang2025depth} utilizes depth-guided bundle sampling. However, these generalizable methods often compromise high-frequency detail for robustness, a limitation our method addresses.

\paragraph{NeRF Super-Resolution.}
NeRF-based SR must recover high-frequency detail while maintaining cross-view consistency, unlike conventional 2D SR~\cite{dong2016srcnn,wang2018esrgan,wang2021realesrgan}.
Prior approaches use HR references~\cite{wang2022nerf,huang2023refsr}, per-scene latent optimization~\cite{han2024super}, or diffusion-based refinement~\cite{lee2024disr}, often incurring additional supervision, optimization cost, or inference overhead.
FastSR-NeRF~\cite{lin2024fastsr} improves efficiency with lightweight SR modules but is less effective at large magnification.
Recent 3DGS SR methods adapt video/2D SR priors~\cite{shen2024supergaussian,wan2025s2gaussiansparseviewsuperresolution3d,chen2025bridging}, but discrete primitives and pretrained priors can introduce overhead and reduce robustness across scenes/resolutions.

\paragraph{Generalizable NeRFs.}
Recent works on generalizable NeRFs aim to reconstruct novel scenes without per-scene optimization. MVSGaussian \cite{mvsgaussian} integrates Gaussian splatting with multi-view stereo for fast reconstruction, but its explicit structure struggles to capture fine-grained HR details. GeFU \cite{gefu} improves quality via geometry-aware fusion and refined rendering, yet still requires scene-specific refinements for consistent HR outputs. GDB-NeRF \cite{fang2025depth} introduces depth-guided bundle sampling for efficient generalization but does not address super-resolution, leaving high-frequency detail underrepresented. Overall, these methods achieve efficiency and generalization but fall short in producing high-resolution, view-consistent outputs directly from LR inputs—a gap our method addresses.

In contrast, we propose Generalizable NGP-SR, which leverages Neural Graphics Primitives (NGP) to reconstruct high-resolution radiance fields directly from LR inputs and camera poses. By learning high-frequency details in the 3D domain, our framework ensures multi-view consistency, generalizes to unseen scenes, and significantly improves both quality and efficiency compared to existing super-resolution and generalizable NeRF approaches.


\section{Generalizable NGP-SR Method}

\begin{figure*}[t!]
    \centering
    \includegraphics[width=1\textwidth]{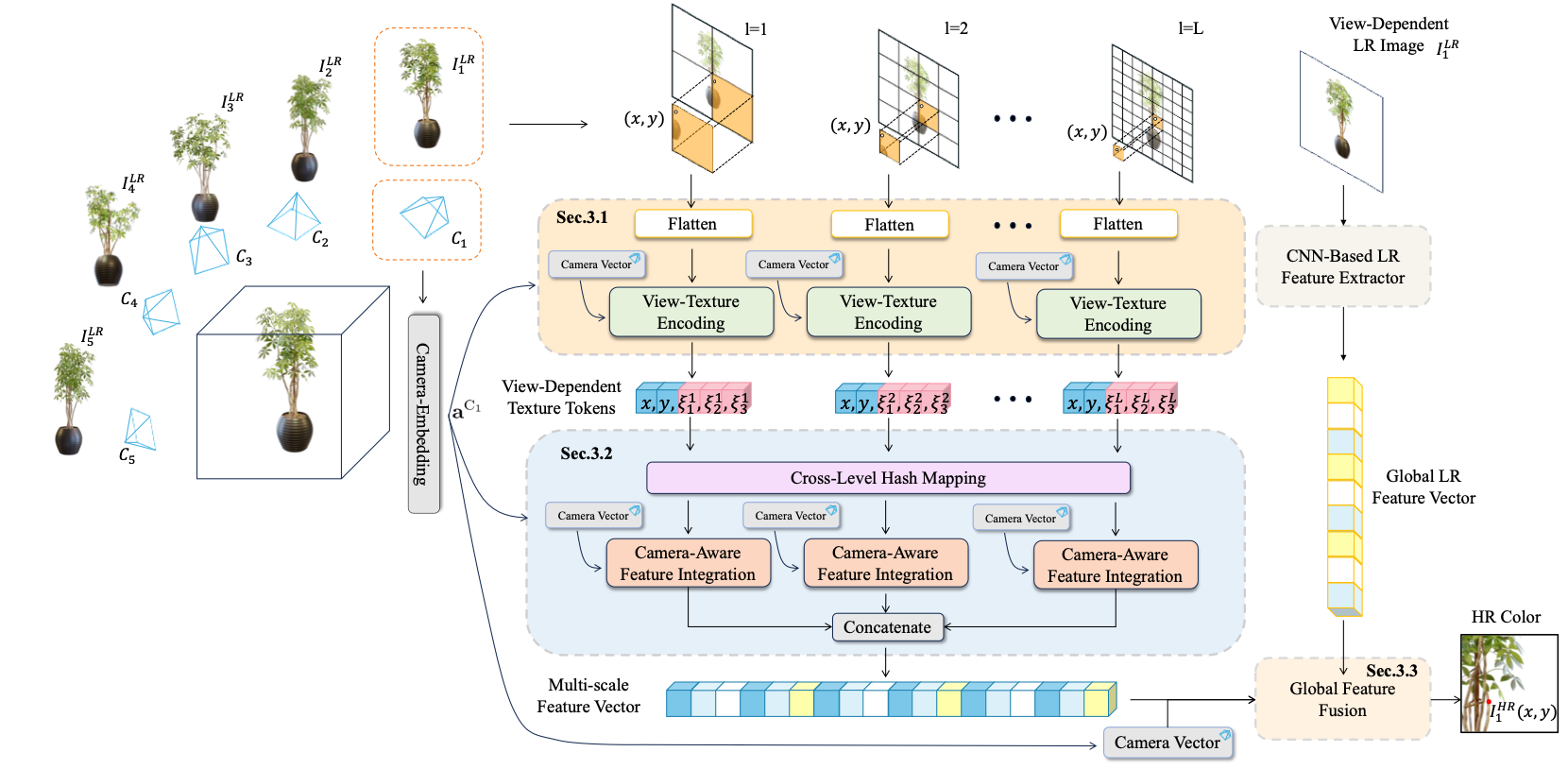}
    \caption{NGP-SR takes low-resolution (LR) images with camera poses and encodes them into hierarchical view-texture tokens that capture local spatial and pose-aware texture cues. These tokens are interpolated through a unified hash table with adaptive, camera-conditioned weights to produce robust multi-resolution feature vectors. Finally, a global feature fusion module integrates holistic LR image context, and a camera-aware decoder synthesizes high-resolution, view-consistent outputs.}
    \label{fig:pipeline}
    \vspace{-10pt}
\end{figure*}

Given a set of low-resolution (LR) images $\{\mathbf{I}^{\text{LR}}_{n}\}^N_{n=1}$ with corresponding camera poses $\{\mathbf{C}_n\}^N_{n=1}$, our goal is to reconstruct a super-resolved radiance field. Once trained, the model can directly synthesize HR outputs for novel viewpoints with camera poses $\{\mathbf{C}'_m\}$ not seen during training, given only their LR renderings.
As illustrated in Fig.~\ref{fig:pipeline}, our framework consists of three key components:
\textbf{1) Hierarchical View-Texture Encoding:} Multi-scale LR image patches and their corresponding camera poses are encoded into fixed-length tokens.
\textbf{2) Camera-Aware Feature Integration:} Camera poses are embedded into the feature generation process, allowing viewpoint-dependent representations that generalize across novel scenes and viewpoints.
\textbf{3) Global Feature Fusion:} We integrate global LR image features to guide the decoding of HR signals, ensuring multi-view consistency and fine-grained detail.

Our design is inspired by VR-INR \cite{aiyetigbo2025implicit}, which employs hierarchical encoding for video SR. Unlike VR-INR, which operates in purely spatial-temporal domains, our method explicitly incorporates 3D geometry through camera-aware integration and feature fusion, enabling generalizable high-resolution novel view synthesis.

\subsection{Hierarchical View-Texture Encoding}

Given LR input images $\mathbf{I}^{\text{LR}}_n$ and their associated camera poses $\mathbf{C}_n$, our goal is to construct compact, view-aware query tokens that serve as the foundation for 3D-consistent super-resolution.

At each resolution level $l$, we extract local patches containing the HR query coordinates:
$
\mathbf{P}_i^l = \mathbf{I}^{\text{LR}}(\mathbf{r}_i^l),
$
where $\mathbf{P}_i^l$ denotes the sampled patch around location $\mathbf{r}_i^l$. These multi-scale patches capture progressively finer spatial context for high-resolution reconstruction, as shown in Fig.~\ref{fig:pipeline}. 

To incorporate viewpoint information, we embed each camera pose $\mathbf{C}_n$ into a latent code via a lightweight MLP:
\begin{equation}
\label{eqn:a_C}
\mathbf{a}^{\text{C}n} = \text{MLP}_a(\mathbf{C}_n),
\end{equation}
which produces a compact representation of the viewing direction. This latent code is injected into the patch encoder $\mathcal{G}_{\text{T}}^l$ to produce pose-aware texture tokens:
\begin{equation}
\label{eqn:g_T}
[\boldsymbol{\xi}^l_1, \boldsymbol{\xi}^l_2, \dots, \boldsymbol{\xi}^l_F] = \mathcal{G}_{\text{T}}^l(\mathbf{P}_i^l, \mathbf{a}^{\text{C}_n}).
\end{equation}

Finally, hierarchical view-texture query tokens are constructed by concatenating spatial coordinates with the corresponding texture tokens:
\begin{equation}
\mathbf{Q}^{l}_{i,n} = [\mathbf{x}, \mathbf{y}, \boldsymbol{\xi}^l_1, \dots, \boldsymbol{\xi}^l_F] = [\mathbf{q}^l_1, \dots, \mathbf{q}^l_S].
\end{equation}

%


\subsection{Camera-Aware Feature Integration}

\begin{figure}[t!]
    \centering
    \includegraphics[width=\columnwidth]{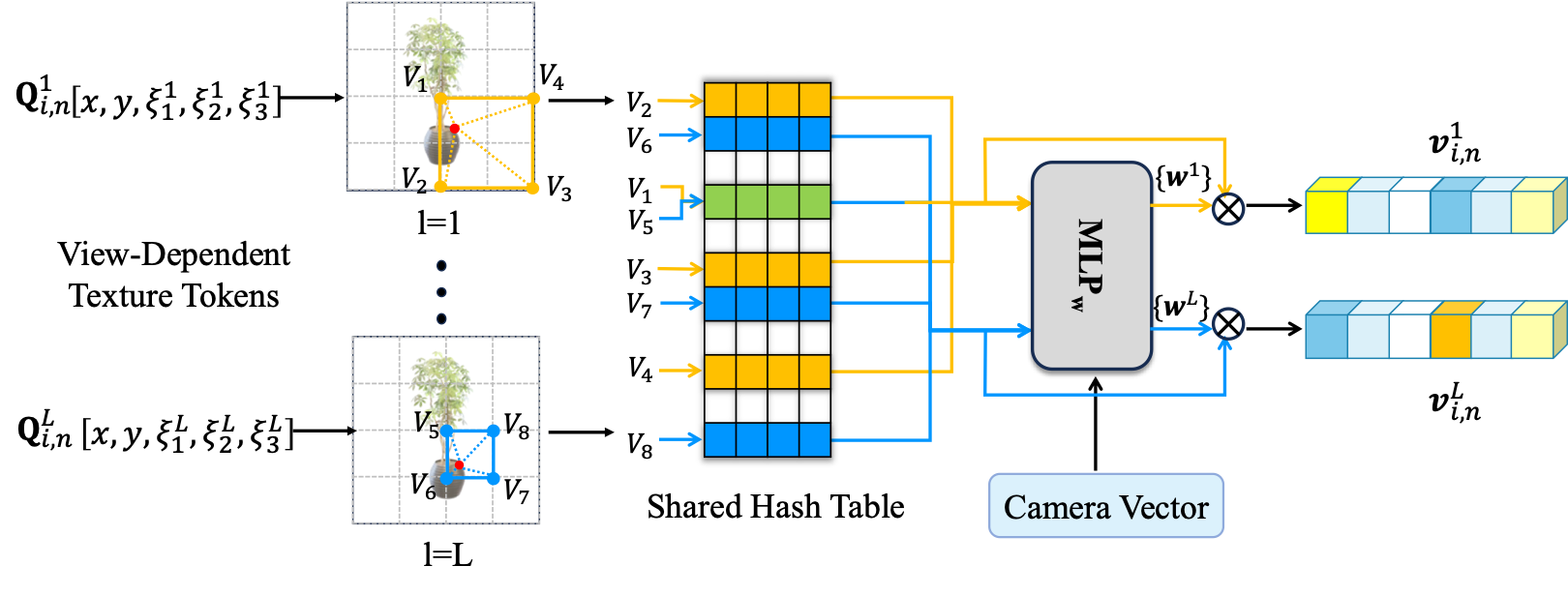}
    \caption{2D illustration of Cross-Level Hash Mapping mechanism. {For each query token, hyper-vertices are identified at multiple resolution levels of the grid. Texture tokens introduce patch-dependent offsets, such that weight of the selected hyper-vertices may differ across resolutions for the same spatial coordinate. To reduce memory consumption, hyper-vertices from coarser levels are mapped to vertices at the finest resolution level.}
}
    \label{fig:mf}
    \vspace{-10pt}
\end{figure}

Given the hierarchical view-texture query tokens $\mathbf{Q}^l_{i,n}$, our goal is to generate point-wise, view-dependent feature vectors that incorporate both multi-scale spatial context and camera information. This is achieved through a unified hashing and implicit interpolation process.

\noindent \textbf{Hyper-Vertex Retrieval.}
For each query token $\mathbf{Q}^l_{i,n}$ at resolution level $l$, we locate its enclosing hyper-vertices $\mathbf{V}^l_{i,n} = \{\mathbf{V}^l_{i,n,k}\}_{k=1}^K$ on the discretized multi-resolution grid, where $K=2^d$ and $d$ is the dimension of $\mathbf{Q}^l{i,n}$. Each hyper-vertex $\mathbf{V}^l_{i,n,k}$ is then mapped to a latent feature vector via a unified hash table:
\begin{equation}
\hat{\mathbf{v}}^l_{i,n,k} = \text{HashTable}(\mathbf{V}^l_{i,n,k}).
\end{equation}

Unlike level-specific hash mappings, our unified design uses a single shared hash table across all $L$ levels, significantly reducing memory usage while preserving multi-scale representation. Fig.~\ref{fig:mf} illustrates such cross-level hash mapping in 2D condition.

\noindent \textbf{Camera-Conditioned Weight Prediction.}
To interpolate the hashed features, we predict adaptive weights conditioned not only on the geometric relation of $\mathbf{Q}^l_{i,n}$ to its surrounding vertices but also on the camera embedding $\mathbf{a}^{C_n}$. Specifically, a lightweight two-layer MLP takes as input the relative offset of $\mathbf{Q}^l_{i,n}$ within its voxel together with $\mathbf{a}^{C_n}$, and outputs interpolation weights:
\begin{equation}
[\mathbf{w}^l_{i,n,1}, \dots, \mathbf{w}^l_{i,n,K}] = \text{MLP}_w\big(\mathbf{Q}^l_{i,n}, \mathbf{a}^{C_n}\big).
\end{equation}

\noindent \textbf{Feature Aggregation.}
The final feature vector corresponding to $\mathbf{Q}^l_{i,n}$ is then computed as the weighted sum of its hyper-vertex features:
\begin{equation}
\mathbf{v}^l_{i,n} = \sum_{k=1}^{K} \mathbf{w}^l_{i,n,k} \cdot \hat{\mathbf{v}}^l_{i,n,k}.
\end{equation}

\noindent \textbf{Multi-Scale Feature Representation.}
Repeating this process across all $L$ resolution levels yields a set of view-dependent feature vectors. Concatenating them forms the final point-wise, multi-scale representation for viewpoint $\mathbf{C}_n$:
\begin{equation}
\mathbf{v}_{i,n} = [\mathbf{v}^1_{i,n}, \mathbf{v}^2_{i,n}, \dots, \mathbf{v}^L_{i,n}].
\end{equation}

\subsection{Global Feature Fusion}
To ensure that high-resolution outputs are both detail-rich and consistent across views, we integrate global low-resolution (LR) image features into the view-dependent feature representations $\mathbf{v}_{i,n}$. This is achieved through a two-stage fusion mechanism that combines viewpoint-conditioned features with scene-level global context.

\noindent \textbf{Stage 1: Camera-Aware Feature Enhancement.}
From the previous step, each query point $\mathbf{Q}_{i,n}$ is represented by its multi-resolution, view-dependent feature vector $\mathbf{v}_{i,n} = [\mathbf{v}^1_{i,n}, \dots, \mathbf{v}^L_{i,n}]$. To explicitly encode viewpoint information, we fuse these features with the camera embedding $\mathbf{a}^{C_n}$. Specifically, the features and embedding are concatenated and passed through a mapping network $\mathcal{G}{\text{cam}}$:
\begin{equation}
\label{eqn:g_cam}
\mathbf{f}_{i,n}^{\text{inter}} = \mathcal{G}_{\text{cam}}(\mathbf{v}_{i,n}, \mathbf{a}^{C_n}),
\end{equation}
yielding an intermediate representation that encodes both local multi-scale textures and viewpoint cues.

\noindent \textbf{Stage 2: Global LR Feature Integration.}
While $\mathbf{f}_{i,n}^{\text{inter}}$ captures local and view-dependent details, it lacks holistic scene context. To address this, we extract global image features from the LR input $\mathbf{I}^{\text{LR}}$ using a CNN-based encoder $\mathcal{G}_{\text{CNN}}$:
\begin{equation}
\label{eqn:g_cnn}
\mathbf{f}_{\text{LR}} = \mathcal{G}_{\text{CNN}}(\mathbf{I}^{\text{LR}}).
\end{equation}
These global features provide structural and semantic guidance, ensuring that reconstructed HR outputs remain coherent across views.

\noindent \textbf{Final Fusion.}
We then combine the intermediate features $\mathbf{f}_{i,n}^{\text{inter}}$ with the global LR features $\mathbf{f}_{\text{LR}}$ through a fusion network $\mathcal{G}_{\text{fuse}}$:
\begin{equation}
\label{eqn:g_fuse}
\bar{\mathbf{v}}_{i,n} = \mathcal{G}_{\text{fuse}}(\mathbf{f}_{i,n}^{\text{inter}}, \mathbf{f}_{\text{LR}}),
\end{equation}
producing global-context aware feature vectors $\bar{\mathbf{v}}_{i,n}$ that guide HR reconstruction with both local and global consistency.

\noindent \textbf{Decoding.}
Finally, the fused feature vectors are decoded into HR color values through a two-layer MLP:
\begin{equation}
\hat{\mathbf{C}}^{\text{HR}}_{i,n} =\text{MLP}_c(\bar{\mathbf{v}}_{i,n}).
\end{equation}
Training is performed by minimizing the pixel-wise Mean Squared Error (MSE) between predictions and ground-truth HR colors:
\begin{equation}
\mathcal{L} = \|\hat{\mathbf{C}}^{\text{HR}}_{i,n} - \mathbf{C}^{\text{HR}}_{i,n}\|_2^2.
\end{equation}
This global fusion strategy ensures that fine details from local patches are preserved, while structural context from the LR image enforces multi-view coherence across synthesized HR outputs.

\section{Experiment}

\subsection{Implementation Details}

Our framework is implemented in PyTorch, implementation requirement depends on the output resolution. We trained on NVIDIA A100 GPUs with a batch size of 8. We adopt the Adam optimizer with an initial learning rate of 0.00125. The model is trained for $\geq$200 epochs/iterations depending on the dataset sizes. Training requires approximately 0.7 hours per scene, while inference achieves 0.7 seconds per image. Detailed implementation requirement will be provided in the supplementary materials.

\begin{table*}
    \caption{Comparison of SOTAs on Blender datasets with varying resolutions and scaling factors. The best results are highlighted in bold, and the second-best results are underlined.}
    \centering
    \resizebox{0.95\textwidth}{!}{%
    \begin{tabular}{lcccccccccccc}
        \toprule
        \multirow{2}{*}{Methods} & \multicolumn{3}{c}{Blender$\times$2 (100 $\times$ 100)} & \multicolumn{3}{c}{Blender$\times$4 (100 $\times$ 100)} & \multicolumn{3}{c}{Blender$\times$2 (200 $\times$ 200)}& \multicolumn{3}{c}{Blender$\times$4 (200 $\times$ 200)}\\
        \cmidrule(lr){2-4} \cmidrule(lr){5-7} \cmidrule(lr){8-10}\cmidrule(lr){11-13}
        & PSNR $\uparrow$ & SSIM $\uparrow$ & LPIPS $\downarrow$ 
        & PSNR $\uparrow$ & SSIM $\uparrow$ & LPIPS $\downarrow$ 
        & PSNR $\uparrow$ & SSIM $\uparrow$ & LPIPS $\downarrow$ 
        & PSNR $\uparrow$ & SSIM $\uparrow$ & LPIPS$\downarrow$ \\
        \midrule
        TensoRF~\cite{chen2022tensorf} &25.41 &0.882 &0.161 &24.06 &0.843 &0.188 &27.27 &0.909 &0.199 &26.18 &0.882 &0.146\\ 
        Instant-NGP~\cite{mueller2022instant} &26.34 &0.916 &0.157 &24.83 &0.872 &0.229 &28.29 &0.934 &0.109&27.07 &0.904 &0.159\\      
        3DGS~\cite{kerbl20233d} &27.57 &0.935 &0.124  &25.59  &0.887 &0.213 &29.49 &{0.949} &0.087&27.86  &0.916 &0.147\\
        LR+Bicubic &28.20 &\textbf{0.946} &0.099 &25.99 &\underline{0.896} &0.192 &30.53 &\underline{0.962} &0.066 &28.60 &\underline{0.927} &0.128\\         
        \midrule
        NeRF-SR~\cite{wang2022nerf} &28.74&0.924&{0.066} &\underline{28.07} &\textbf{0.921} &\underline{0.071} &30.82 &0.953 &{0.055}&28.46 &0.921 &0.076\\        
        FastSR-NeRF~\cite{lin2024fastsr} &27.27 &0.902 &0.142 &22.32 &0.453 &0.475 &31.42 &{0.952} &0.047 &\textbf{30.47} &\textbf{0.944} &0.075\\
        SuperGaussian~\cite{shen2024supergaussian} &26.39 &0.915 &0.090 & -- &-- &--  &29.91 &0.947 &\underline{0.035}&28.44 &{0.923} &\underline{0.067}\\
        S2Gaussian~\cite{wan2025s2gaussiansparseviewsuperresolution3d} &--  & -- & - & -- &-- &--  & -- & -- & -- &24.19 &{0.879} & {0.089}\\
        \midrule
        TensoRF+Ours &26.61 &0.908 &0.093 &23.52 &0.863 &0.119 &28.28 &0.921 &0.068 &26.10 &0.891 &0.103\\        
        Instant-NGP+Ours &29.25 &\underline{0.945} &0.062 &25.13 &0.888 &0.098 &31.02 &{0.952} &0.041 &27.17 &0.892 &{0.085}\\
        3DGS+Ours &\textbf{29.43} &\textbf{0.946} &\underline{0.051} &{25.80} &0.893 &0.085 &\underline{32.17} &0.940 &0.041 &27.14 &0.892 &0.076\\
        LR+Ours &\underline{29.34} &0.938 &\textbf{0.049} &\textbf{28.77} &\textbf{0.921} &\textbf{0.069} &\textbf{33.68} &\textbf{0.971} &\textbf{0.027} &\underline{29.31} &0.923 &\textbf{0.055}\\        
        \bottomrule
    \end{tabular}
    }
    \label{blender-comparison}
\end{table*}

\begin{table*}
    \caption{Comparison of SOTAs on LLFF datasets with varying resolutions and scaling factors. The best results are highlighted in bold, and the second-best results are underlined.}
    \centering
    \resizebox{0.95\textwidth}{!}{%
    \begin{tabular}{lcccccccccccc}
        \toprule
        \multirow{2}{*}{Methods} & \multicolumn{3}{c}{LLFF$\times$2 (100 $\times$ 100)} & \multicolumn{3}{c}{LLFF$\times$4 (100 $\times$ 100)} & \multicolumn{3}{c}{LLFF$\times$2 (200 $\times$ 200)}  & \multicolumn{3}{c}{LLFF$\times$4 (200 $\times$ 200)}\\
        \cmidrule(lr){2-4} \cmidrule(lr){5-7} \cmidrule(lr){8-10}\cmidrule(lr){11-13} 
        & PSNR $\uparrow$ & SSIM $\uparrow$ & LPIPS $\downarrow$ & PSNR $\uparrow$ & SSIM $\uparrow$ & LPIPS $\downarrow$ & PSNR $\uparrow$ & SSIM $\uparrow$ & LPIPS $\downarrow$ & PSNR $\uparrow$ & SSIM $\uparrow$ & LPIPS $\downarrow$\\
        \midrule
        TensoRF~\cite{chen2022tensorf} &24.07 &0.754 &0.314 &22.16 &0.591 &0.596 &24.69 &0.750 &0.309 &23.49 &0.648 &0.503 \\  
        Instant-NGP~\cite{mueller2022instant} &22.17 &0.667 &0.335 &20.90 &0.543 &0.479 &21.47 &0.616 &0.332 &20.81 &0.571 &0.428 \\     
        3DGS~\cite{kerbl20233d} &23.20 &0.735 &0.307 &21.52 &0.581 &0.580 &23.76 &0.748 &0.294 &22.73 &0.654 &0.478\\  
        LR+Bicubic &24.56 &0.765 &0.299 &22.62 &0.608 &0.579 &25.92 &0.793 &0.260 &24.43 &0.687 &0.468\\         
        \midrule
        NeRF-SR~\cite{wang2022nerf} &{26.24} &{0.838} &\underline{0.123} &24.07 &0.713 &{0.252} &\underline{27.15} &{0.837} &\underline{0.122} &{25.44} &{0.745} &{0.208}\\
        FastSR-NeRF~\cite{lin2024fastsr} &24.33 &0.765 &0.297 &21.20 &0.584 &0.473 &26.21 &0.822 &0.241 &21.30 &0.384 &0.475 \\
        3DSR ~\cite{chen2025bridging} &26.38  & \underline{0.886} & 0.181 &24.18  &\underline{0.760} &0.342  & 24.90 & \underline{0.838}  & 0.235  & \underline{25.88} & \underline{0.809} & 0.261\\
        S2Gaussian~\cite{wan2025s2gaussiansparseviewsuperresolution3d} &--  & -- & - & -- &-- &--  & -- & -- & -- &20.45 &{0.654} & \underline{0.139}\\
        \midrule
        TensoRF+Ours &\underline{26.46} &{0.805} &0.181 &\underline{25.34} &{0.751} &0.190 &{27.07} &0.809 &0.170 &24.49 &0.718 &0.213\\   Instant-NGP+Ours &23.31 &0.719 &0.171 &22.30 &0.689 &0.206 &21.74 &0.646 &0.155 &20.92 &0.588 &0.264\\
        3DGS+Ours &26.11 &{0.805} &{0.168} &{25.07} &{0.754} &\underline{0.183} &26.87 &{0.828} &{0.149} &{25.42} &{0.749} &{0.176}\\
        LR+Ours &\textbf{30.32} &\textbf{0.892} &\textbf{0.072} &\textbf{29.29} &\textbf{0.847} &\textbf{0.112} &\textbf{32.90} &\textbf{0.934} &\textbf{0.040} &\textbf{29.61} &\textbf{0.874} &\textbf{0.074}\\
        \bottomrule
    \end{tabular}
    }
    \label{llff-comparison}
\end{table*}
\begin{figure*}[t!]
    \centering
    \includegraphics[width=0.8\textwidth]{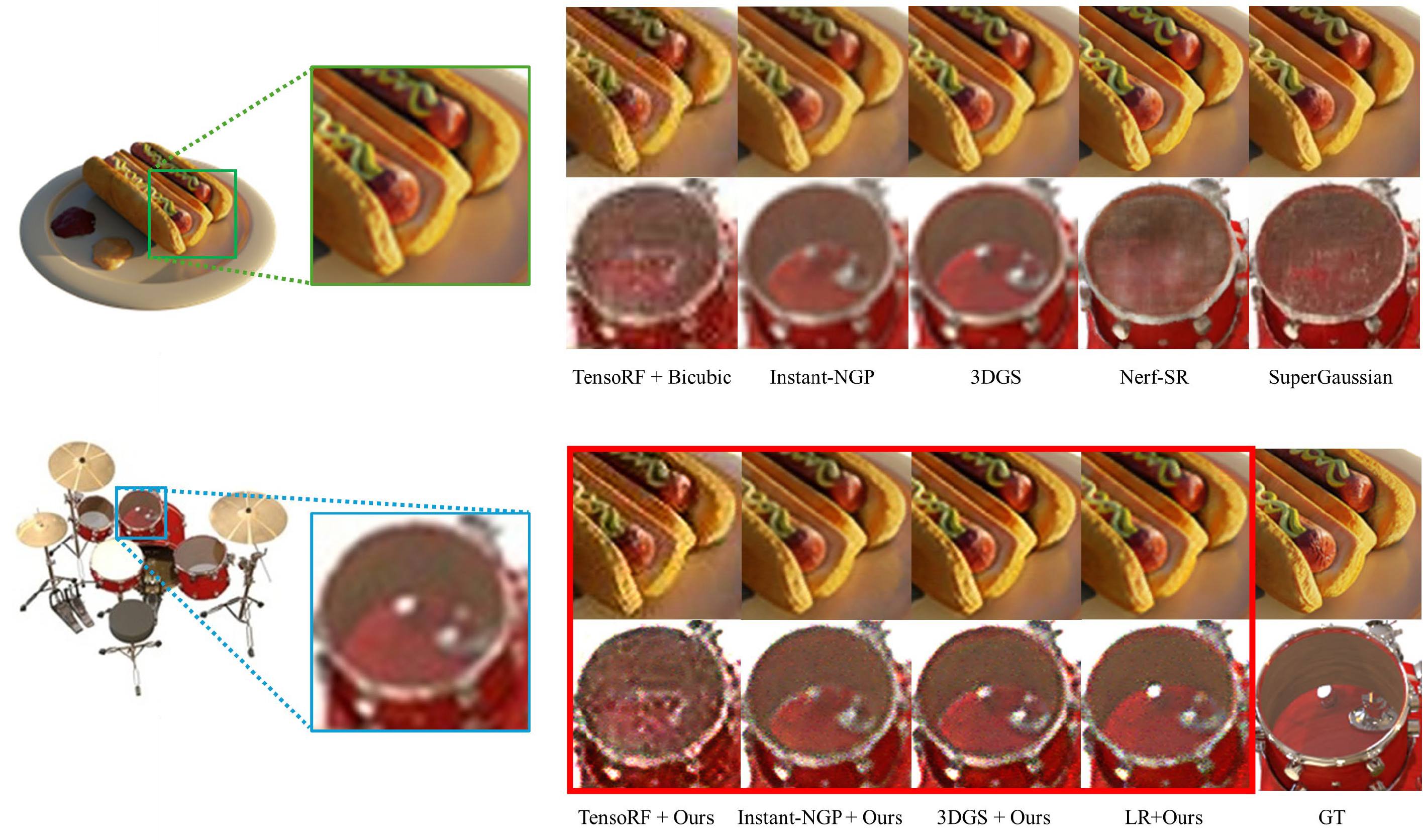}
    \caption{The qualitative comparison of our method on the Blender datasets, with input images at 200$\times$200 resolution and $\times$4 super-resolution. The leftmost images show the low-resolution inputs, while the images highlighted in red correspond to our results. Compared with other methods that fail to recover fine structures, our approach successfully reconstructs sharp and detailed results. Notably, our method consistently enhances regions where other approaches produce blurred or low-frequency reconstructions.}
    \label{blender}
\end{figure*}

\begin{figure*}[t!]
    \centering
    \includegraphics[width=0.8\textwidth]{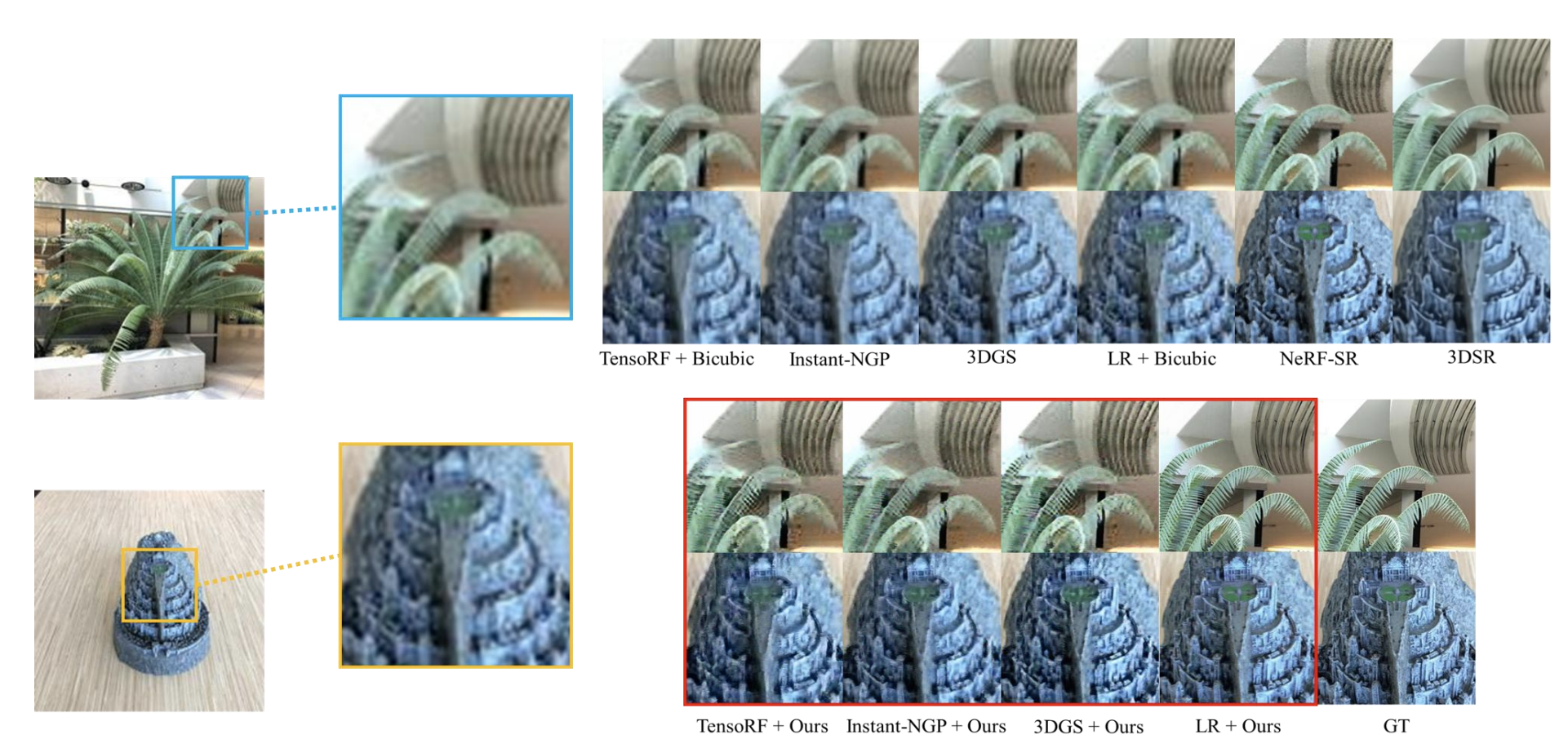}
    \caption{The qualitative comparison of our method on the LLFF datasets, with input images at 200$\times$200 resolution and $\times$4 super-resolution.The leftmost images show the low-resolution inputs, while the images highlighted in red correspond to our results. Our method successfully reconstructs fine details in leaves and architectural structures, and consistently improves reconstruction quality in regions where other methods exhibit degraded results.
}
    \label{llff}
\end{figure*}

\vspace{5pt}
\noindent \textbf{Camera Pose Embedding ($\mathbf{a}^{\text{C}_n}$, Eqn.~\ref{eqn:a_C}).}
We encode each camera pose $\mathbf{C}_n$ into a compact representation using a three-layer MLP. The first two layers have 64 neurons with ReLU activations, followed by a final 16-dimensional output layer with Tanh activation to ensure bounded values in $[-1,1]$.

\noindent \textbf{Texture Encoder Network ($\mathcal{G}_{\text{T}}^l$, Eqn.~\ref{eqn:g_T}).}
For each resolution level $l$, the texture encoder is a two-layer MLP that processes vectorized LR image patches. Each hidden layer contains 128 neurons with ReLU activation, and the output is a compact texture code of dimension $F=3$.

\noindent \textbf{Camera-Aware Mapping Network ($\mathcal{G}_{\text{cam}}$, Eqn.~\ref{eqn:g_cam}).}
To integrate viewpoint information, we concatenate the multi-resolution feature vector $\mathbf{v}_{i,n}$ with the camera embedding $\mathbf{a}^{C_n}$, and feed the result into a four-layer MLP. Each layer has 64 neurons with Tanh activation. The hidden layers internally allocate 32 dimensions to camera embedding features, ensuring effective fusion between spatial and viewpoint cues.

\noindent \textbf{Global Feature Extractor ($\mathcal{G}_{\text{CNN}}$, Eqn.~\ref{eqn:g_cnn}).}
Global context is captured using a lightweight CNN with three convolutional layers ([64, 64, 32] channels), $3 \times 3$ kernels, and ReLU activations. An adaptive average pooling layer produces a 32-dimensional global feature vector.

\noindent \textbf{Two-Stage Fusion Network ($\mathcal{G}_{\text{fuse}}$, Eqn.~\ref{eqn:g_fuse}).}
The fusion module first combines hierarchical features and camera embeddings via a four-layer MLP (64 neurons, Tanh activations). The intermediate fused representation is then merged with global LR features in a second four-layer MLP of the same configuration. Both stages use intermediate embeddings of size 32 for efficient integration of local and global information.
For clarity and completeness, additional implementation details are provided in the supplementary materials.

\subsection{Comparison Experiments}

\noindent \textbf{Benchmarks and Protocol.}
We evaluate our approach on three widely used datasets for novel view synthesis (NVS) and super-resolution.
\textbf{Blender}\cite{mildenhall2020nerf}: 8 synthetic scenes with diverse shapes and materials. We train on the training sets and evaluate on the held-out, unseen test sets of all eight scenes. Input images are downsampled to $100{\times}100$ and $200{\times}200$, and results are reported for SR scales of $\times2$ and $\times4$.
\textbf{LLFF}\cite{mildenhall2019llff}: eight forward-facing real-world scenes. Following NeRF-SR~\cite{wang2022nerf}, we train on the entire dataset due to the limited number of images per scene. Evaluation is conducted on SR novel views synthesized by different NeRF backbones.
\textbf{DTU}~\cite{dtu}: a multi-view stereo benchmark with 124 indoor scenes captured under seven lighting conditions. We use 88 training scenes with $128{\times}128$ LR inputs and $\times4$ SR, and evaluate on 8 held-out DTU test set.
\noindent \textbf{Metrics.}
We employ standard metrics, including Peak Signal-to-Noise Ratio (PSNR), Structural Similarity Index Measure (SSIM), and Learned Perceptual Image Patch Similarity (LPIPS) \cite{zhang2018unreasonable}, to comprehensively assess pixel accuracy, structural fidelity, and perceptual quality of SR results. For methods without SR, their outputs are bicubically upsampled to match target resolution. We also report \textit{LR+Ours} (downsampled HR ground-truth as LR input, super-resolved with ours) to indicate the attainable upper bound of our framework.

\vspace{5pt}
\noindent \textbf{Compared Methods.}
We group baselines into three categories:
1) \emph{Backbone-only NVS without SR}: TensoRF~\cite{chen2022tensorf}, Instant-NGP~\cite{mueller2022instant}, and 3D Gaussian Splatting (3DGS)~\cite{kerbl20233d}. Since these methods do not perform SR, we upsample their LR outputs using bicubic interpolation ($\times2$, $\times4$).
2) \emph{NeRF-based SR}: NeRF-SR~\cite{wang2022nerf} and FastSR-NeRF~\cite{lin2024fastsr}. Both require HR reference supervision, which limits their applicability in low-resolution only settings. As FastSR-NeRF is not open-sourced, we directly report the numbers from their paper; hence, no visual comparison is available.
3) \emph{3DGS + video SR}: SuperGaussian~\cite{shen2024supergaussian}, which adapts pretrained video SR models to upsample continuous 3DGS renderings. Since results on the LLFF dataset are not provided, we only compare against SuperGaussian on Blender (Tab.~\ref{blender-comparison}). For S2Gaussian~\cite{wan2025s2gaussiansparseviewsuperresolution3d}, since the authors do not provide publicly available code, we directly report the results as presented in the original paper. And 3DSR~\cite{chen2025bridging} on LLFF dataset.
4) \emph{Generalizable NeRFs}: MVSGaussian~\cite{mvsgaussian}, GeFU~\cite{gefu}, and GDB-NeRF~\cite{fang2025depth}, evaluated on DTU with input images resized to $512{\times}512$. 

\begin{figure}[t]
    \centering
    \includegraphics[width=\columnwidth]{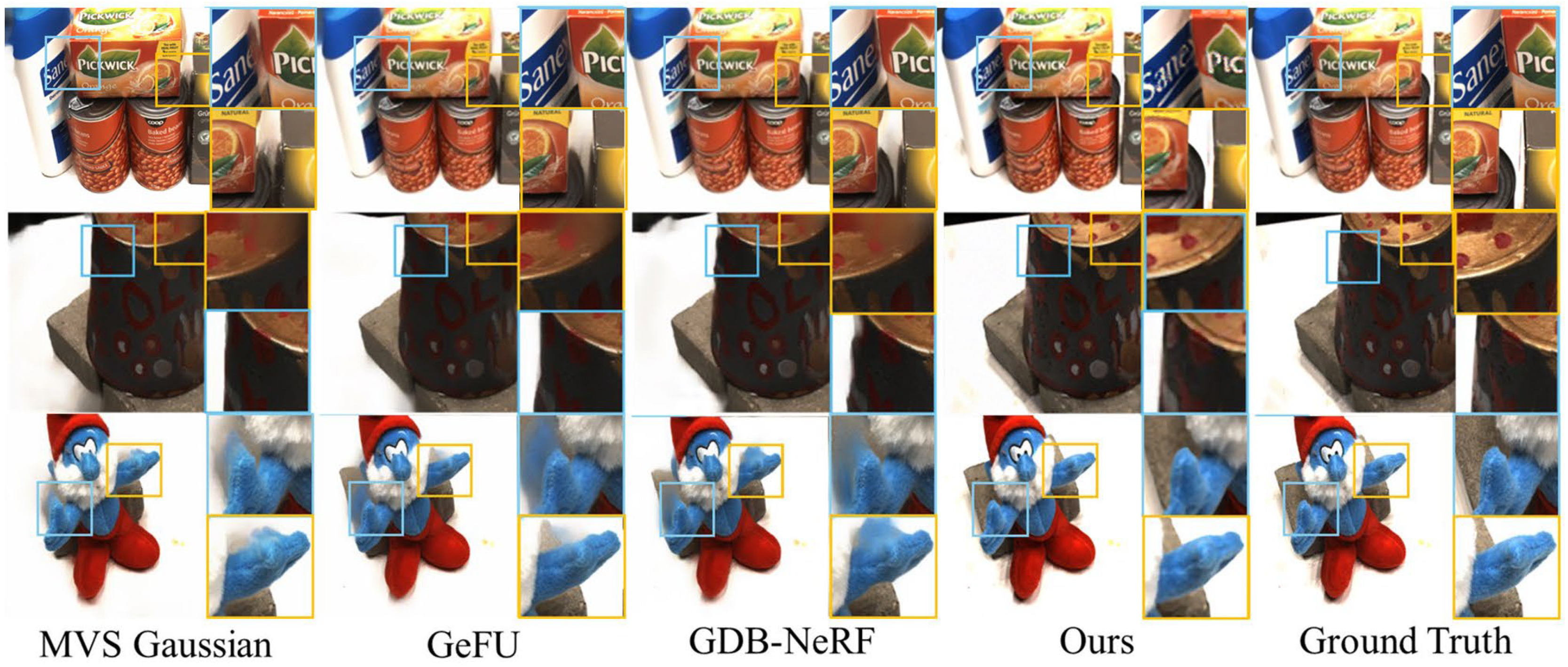}
    \caption{Qualitative evaluation of novel-scene generalization on the DTU dataset. Each panel shows the reconstructed image (left) and two zoomed-in crops (right). {Baselines use $512{\times}512$ inputs, while our method operates on $128{\times}128$ inputs with $\times4$ super-resolution and still reconstructs sharp details with accurate lighting.}}
    \label{generalization}
    \vspace{-10pt}
\end{figure}

\vspace{5pt}
\noindent \textbf{Quantitative and Qualitative Analysis.}
Tables~\ref{blender-comparison}, \ref{llff-comparison}, and \ref{DTU-comparison} summarize quantitative results, and Figures~\ref{generalization},\ref{blender} and \ref{llff}, and  provide visual comparisons. Backbone-only NVS methods degrade severely at extreme low-resolution inputs ($100{\times}100$ and $200{\times}200$), in many cases performing worse than bicubic upsampling. This is due to the reduced number of rays and the strong ambiguity of LR observations, which prevent accurate high-frequency reconstruction. In contrast, when combined with our NGP-SR, NeRF and 3DGS backbones achieve significant improvements across all metrics, demonstrating the effectiveness of our framework as a modular SR component.
On Blender, our method outperforms both NeRF-SR and FastSR-NeRF at extremely low resolutions ($100{\times}100$, despite their reliance on HR references, while also surpassing SuperGaussian and in most cases. On LLFF, NGP-SR achieves especially strong performance, which can be attributed to training on all images per scene—similar to NeRF-SR, and the tendency of our NGP-based interpolation to overfit more effectively to observed views. This property allows NGP-SR to reconstruct high-fidelity details when test views share similar content with the training set.
On DTU, NGP-SR generalizes well to unseen test scenes, producing sharper details and more consistent reconstructions from LR inputs and camera poses.
\begin{figure}[!htb]  
  \centering
  \includegraphics[width=\columnwidth]{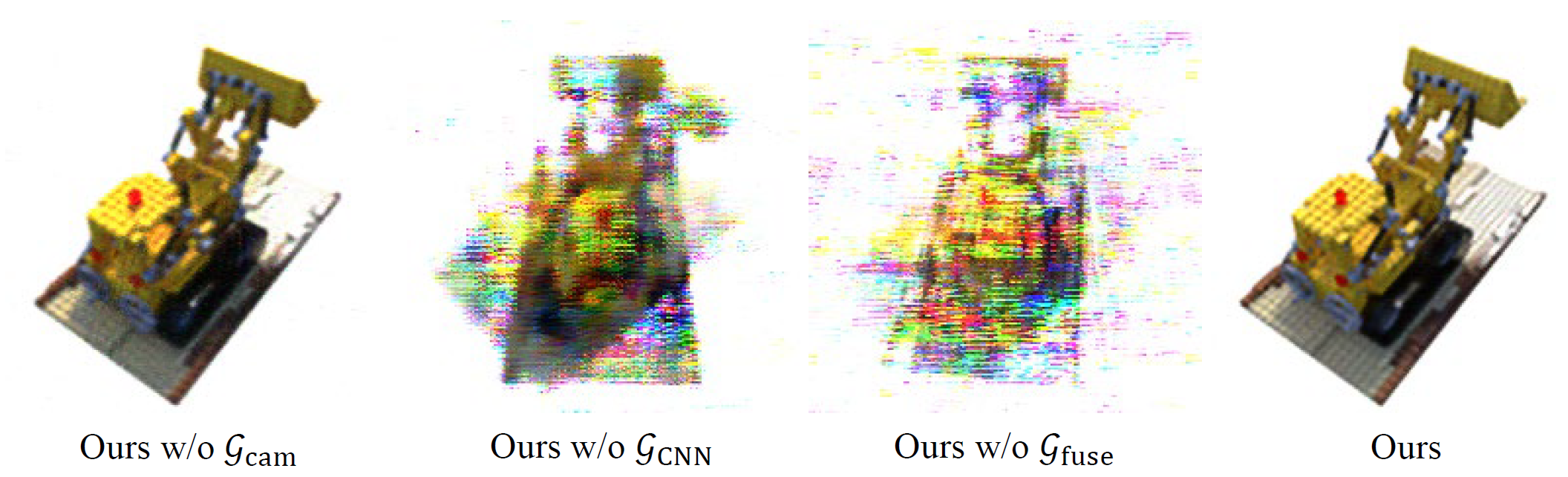}
  \caption{Quantitative ablation on the Lego scene.}
  \label{fig:ablation}
\end{figure}

\begin{table}
    \centering
    \small
    \caption{Comparison of SOTAs on DTU datasets with $512{\times}512$ resolution.}    
    \begin{tabular}{lccc}
        \toprule
        \multirow{2}{*}{Methods} & \multicolumn{3}{c}{DTU}\\
        \cmidrule(lr){2-4}  
        & PSNR $\uparrow$ & SSIM $\uparrow$ & LPIPS$\downarrow$\\
        \midrule
       MVSGaussian &21.95 &0.863 &0.112 \\
        GeFU &23.15 &0.880 &\textbf{0.100} \\
        GDB-NeRF &22.42 &0.867 &0.111 \\
        Ours ($\times$4) &\textbf{31.38} &\textbf{0.884} &0.122 \\
        \bottomrule
    \end{tabular}
    \label{DTU-comparison}
\end{table}

\section{Ablation Study}
\begin{table}
    \centering
    \small
    \caption{Ablation studies of our method on the Lego dataset using PSNR, SSIM, and LPIPS metrics.}    
    \begin{tabular}{lcccc}
        \toprule
        \multirow{2}{*}{Methods} & \multicolumn{3}{c}{Lego$\times$2 (100 $\times$ 100)} \\
        \cmidrule(lr){2-4}
        & PSNR $\uparrow$ & SSIM $\uparrow$ & LPIPS $\downarrow$\\
        \midrule
        Ours w/o $\mathcal{G}_{\text{cam}}$ &26.69&0.892&0.082\\
        Ours w/o $\mathcal{G}_{\text{CNN}}$ &13.38&0.538&0.471\\
        Ours w/o $\mathcal{G}_{\text{fuse}}$ &13.69&0.585&0.436\\
        Ours &\textbf{30.60}&\textbf{0.952}&\textbf{0.075}\\
        \bottomrule
    \end{tabular}

    \label{ablation-lego}
\end{table}

We conduct comprehensive ablation studies to analyze the contributions of key components in our framework.
To systematically assess the effectiveness of these modules, we create subnetworks by individually removing each component and report both qualitative and quantitative outcomes on the Lego scene from the Blender dataset, using a resolution scale factor of $\times2$ and input resolution of $100 \times 100$. 
Results are summarized in Table \ref{ablation-lego} and qualitatively illustrated in Figure \ref{fig:ablation}.

\vspace{5pt}
\noindent \textbf{Camera-Aware Mapping Network ($\mathcal{G}_{\text{cam}}$)} This module integrates view-dependent feature vectors with camera pose embeddings via an MLP, enabling robust viewpoint adaptation. For ablation, we replace $\mathcal{G}_{\text{cam}}$ with a simpler two-layer MLP without camera pose conditioning. The resulting model suffers a notable performance drop (PSNR: -3.27 dB, SSIM: -0.053, LPIPS: +0.02), confirming the critical role of explicit camera pose embedding.

\vspace{5pt}
\noindent \textbf{Global Feature Extractor ($\mathcal{G}_{\text{CNN}}$)} This component captures structural and textural context from low-resolution inputs. Ablating this module significantly degrades reconstruction quality (PSNR: -16.58 dB, SSIM: -0.407, LPIPS: +0.409), demonstrating its importance in providing global context. Qualitative results in Figure \ref{fig:ablation} highlight clear failures in texture and pose reconstruction when $\mathcal{G}_{\text{CNN}}$ is omitted.

\vspace{5pt}
\noindent \textbf{Two-Stage Fusion Network ($\mathcal{G}_{\text{fuse}}$)} This network fuses local view-dependent features with global contextual information in a two-stage MLP process. Removing this fusion step and directly feeding hierarchical features into the decoder dramatically worsens performance (PSNR: -16.27 dB, SSIM: -0.360, LPIPS: +0.374), indicating its crucial role in integrating detailed local and global context. Visual outputs without this fusion module exhibit severe deficiencies in both structure and texture (Figure \ref{fig:ablation}).

\section{Conclusion}
We presented NGP-SR, a generalizable super-resolution framework for Neural Radiance Fields that reconstructs high-resolution views directly from low-resolution multi-view inputs. By combining hierarchical spatial-texture encoding, camera-aware feature integration, and global feature fusion, NGP-SR produces detail-preserving and view-consistent reconstructions.
Experiments on Blender, LLFF, and DTU show that NGP-SR outperforms state-of-the-art methods, especially in challenging low-resolution settings where standard NeRF and 3DGS backbones struggle. Our framework generalizes well to unseen scenes without per-scene retraining, and ablation studies confirm the importance of each component. Our approach reconstructs high-frequency details from LR inputs, without any generative models or high-frequency cues, high-frequency signals may be challengable despite visually sharp reconstructions. Future work will extend NGP-SR to high-frequency signals, dynamic scenes and explore lighter encodings for real-time deployment.

{
    \small
    \bibliographystyle{ieeenat_fullname}
    \bibliography{main}
}


\end{document}